\documentclass[letterpaper]{article} 
\usepackage{aaai2026}  
\usepackage{times}  
\usepackage{booktabs}
\usepackage{multirow}
\usepackage{makecell}
\usepackage{helvet}  
\usepackage{courier}  
\usepackage[hyphens]{url}  
\usepackage{graphicx} 
\usepackage{natbib}  
\usepackage{caption} 
\usepackage{algorithm}
\usepackage{algcompatible}
\usepackage{algpseudocode}
\usepackage{amsmath}
\usepackage{amsfonts}
\usepackage{booktabs}
\usepackage{multirow}
\usepackage{makecell}
\usepackage{subcaption}
\usepackage{tabularx}
\usepackage[table]{xcolor} 
\usepackage{amsthm}
\usepackage{appendix}
\usepackage[table,dvipsnames]{xcolor}
\usepackage{xcolor} 
\usepackage{graphicx}
\usepackage{color}
\usepackage{colortbl}   
\usepackage{url}   
\usepackage{pdfpages}

\definecolor{darkgreen}{RGB}{0, 100, 0}
\definecolor{purple}{RGB}{128, 0, 128}

\usepackage{newfloat}
\usepackage{float}
\usepackage{listings}
\usepackage{mathtools}
\newtheorem{definition}{Definition}
\newtheorem{proposition}{Proposition}

\theoremstyle{plain}

\theoremstyle{definition}

\theoremstyle{remark}

\usepackage{newfloat}
\usepackage{listings}
\DeclareCaptionStyle{ruled}{labelfont=normalfont,labelsep=colon,strut=off} 
\floatstyle{ruled}
\newfloat{listing}{tb}{lst}{}
\floatname{listing}{Listing}
\title{T-Retriever: Tree-based Hierarchical Retrieval Augmented Generation for Textual Graphs}
\author{
    Chunyu Wei\textsuperscript{\rm 1}\equalcontrib,
    Huaiyu Qin\textsuperscript{\rm 1}\equalcontrib,
    Siyuan He\textsuperscript{\rm 1},
    Yunhai Wang\textsuperscript{\rm 1},
    Yueguo Chen\textsuperscript{\rm 1}\thanks{Corresponding author. He works at BRAIN of RUC.}
}
\affiliations{
    \textsuperscript{\rm 1}Renmin University of China, China\\
    weicy15@icloud.com
}

\begin{document}

\maketitle

\begin{abstract}
Retrieval-Augmented Generation (RAG) has significantly enhanced Large Language Models' ability to access external knowledge, yet current graph-based RAG approaches face two critical limitations in managing hierarchical information: they impose rigid layer-specific compression quotas that damage local graph structures, and they prioritize topological structure while neglecting semantic content. We introduce T-Retriever, a novel framework that reformulates attributed graph retrieval as tree-based retrieval using a semantic and structure-guided encoding tree. Our approach features two key innovations: (1) Adaptive Compression Encoding, which replaces artificial compression quotas with a global optimization strategy that preserves the graph's natural hierarchical organization, and (2) Semantic-Structural Entropy (S²-Entropy), which jointly optimizes for both structural cohesion and semantic consistency when creating hierarchical partitions. Experiments across diverse graph reasoning benchmarks demonstrate that T-Retriever significantly outperforms state-of-the-art RAG methods, providing more coherent and contextually relevant responses to complex queries.
\end{abstract}

%
\begin{links}
\link{Code}{https://github.com/T-Retriever/T-Retriever}
\end{links}

\section{Introduction}
\label{sec:introduction}

Large Language Models (LLMs) have revolutionized artificial intelligence~\cite{brown2020language, touvron2023llama}, yet they still struggle with complex structured data. A significant portion of real-world information—from scientific knowledge to social networks and enterprise data—naturally exists as \textbf{attributed graphs}~\cite{DBLP:conf/www/WeiBBW22,DBLP:conf/kdd/WeiHHWCB025,DBLP:journals/aei/ZhangYFZYW25}. Enabling LLMs to effectively reason over these structures is crucial for unlocking deeper insights~\cite{he2024g,DBLP:conf/icann/ZhangWYFJ24}. In response, Graph-based Retrieval-Augmented Generation (RAG) approaches~\cite{edge2024local, peng2024graph} have emerged to connect LLMs with external knowledge captured in graphs.

A critical limitation in current graph-based RAG approaches is the absence of sophisticated hierarchical knowledge management for effective multi-resolution context retrieval, as illustrated in Figure~\ref{fig:illustration}. Effectively managing large attributed graphs requires partitioning them into hierarchies that organize information at different granularity levels. Current systems like GraphRAG~\cite{edge2024local} and HiRAG~\cite{huang2025retrieval} typically rely on conventional community detection algorithms such as Leiden~\cite{traag2019louvain}, which present two fundamental limitations when building \textit{hierarchical} indices:

\begin{figure}[t]
    \centering
    \includegraphics[width=0.92\columnwidth]{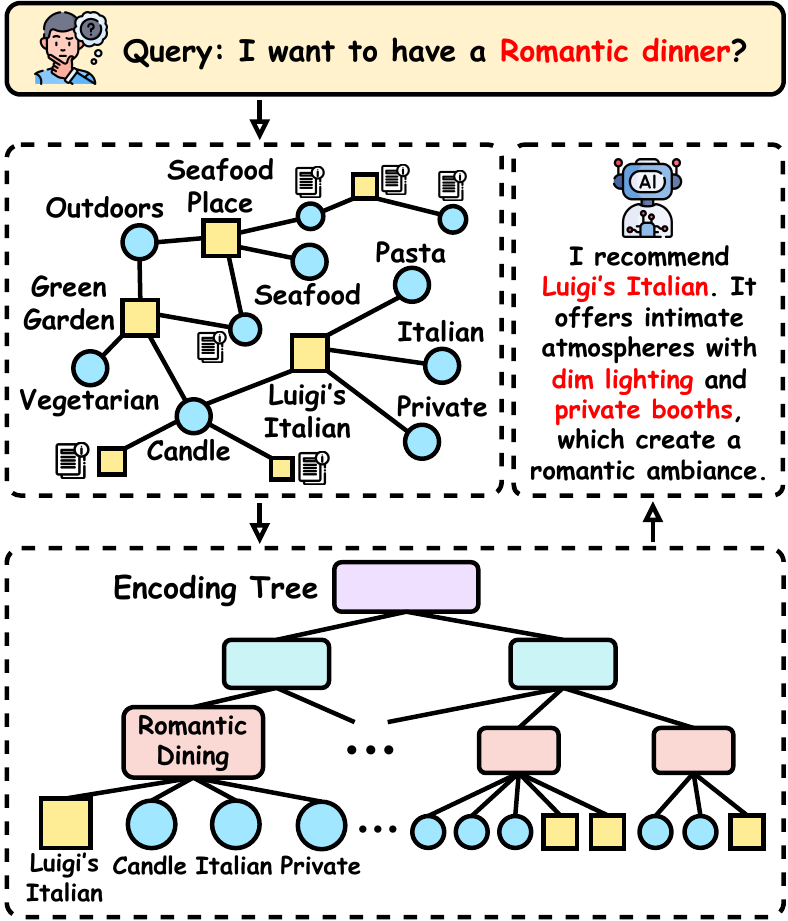}
    \caption{Illustration of \texttt{T-Retriever}. Hierarchical organization of attributed graph knowledge enabling effective multi-resolution context retrieval for question answering.}
    \label{fig:illustration}
\end{figure}

\begin{itemize}
    \item \textbf{Suboptimal Hierarchical Partitioning:} Conventional algorithms impose rigid, predefined layer-specific compression quotas that damage local graph structures and fail to adapt to the data's intrinsic organization. Their bottom-up construction treats layers in isolation, hindering semantic coherence across hierarchical levels and producing representations misaligned with the graph's natural multi-resolution structure.
    
    \item \textbf{Semantic-Structural Disconnect:} These methods predominantly focus on topological structure while neglecting the rich semantic information embedded in node and edge attributes. This oversight results in hierarchical clusters that may be structurally coherent but semantically inconsistent, severely limiting the RAG system's ability to synthesize knowledge requiring both structural and semantic understanding.
\end{itemize}

To address these limitations, we propose a paradigm shift from graph-based to \textbf{tree-based retrieval} with \textbf{\texttt{T-Retriever}}, a system employing a semantic and structure-guided encoding tree for hierarchical representation and retrieval. \texttt{T-Retriever} features two innovations:

First, \textbf{Adaptive Compression Encoding} overcomes suboptimal partitioning through a top-down approach inspired by Shannon-Fano coding~\cite{connell1973huffman}. This replaces rigid quotas with a global optimization strategy that recursively divides the graph based on joint entropy. Unlike bottom-up methods that can disrupt semantic coherence, our approach preserves local structural patterns and cross-layer dependencies while maintaining a global perspective that better reflects the graph's intrinsic multi-resolution structure.

Second, \textbf{Semantic-Structural Entropy} (S²-Entropy) addresses the semantic-structural disconnect by quantifying information via the joint distribution of graph topology and attribute semantics. Minimizing S²-Entropy during encoding ensures the construction of clusters that are both structurally cohesive and semantically consistent, enabling more effective retrieval and reasoning.

Our main contributions are:
\begin{itemize}
    \item We reformulate graph retrieval as tree-based retrieval, proposing Adaptive Compression Encoding to preserve the graph's natural hierarchical organization.
    \item We introduce Semantic-Structural Entropy, integrating both structural patterns and semantic content to guide the construction of a unified, semantically and structurally coherent encoding tree.
    \item Experiments on graph reasoning benchmarks demonstrate that \texttt{T-Retriever} significantly outperforms state-of-the-art RAG methods in graph-related scenarios.
\end{itemize}
\section{Related Work}
\label{sec:relatedwork}
\paragraph{Graph-based Retrieval for Large Language Models. }
The synergy between large language models (LLMs) and knowledge graphs (KGs) is crucial for enhancing reasoning and mitigating hallucination \cite{pan2024unifying, clark2019does}. While early methods required costly model fine-tuning \cite{sun2019ernie, yasunaga2021qa}, the advent of Retrieval-Augmented Generation (RAG) \cite{lewis2020retrieval} has shifted focus to in-context learning.

This paradigm was extended to graphs, creating the field of GraphRAG \cite{gao2023retrieval, sen2023knowledge}. Unlike standard RAG, GraphRAG retrieves interconnected nodes from a graph, providing structured context to the LLM. Foundational methods like G-Retriever \cite{he2024g} established the efficacy of this approach by textualizing retrieved subgraphs. Our work builds on this paradigm by optimizing the core retrieval step.

\paragraph{Hierarchical Indexing for RAG}
To handle large-scale data, hierarchical indexing has become critical. In the textual domain, RAPTOR \cite{sarthi2024raptor} builds a hierarchy via bottom-up clustering and summarization. This concept has been extended to graphs: ArchRAG \cite{wang2025archrag} applies community detection, while HippoRAG \cite{jimenez2024hipporag} uses a Personalized PageRank (PPR) traversal at query time. These methods represent a significant step forward, but their hierarchy construction often relies on heuristics or costly online computation. In contrast, our approach builds the hierarchy offline using a principled, information-theoretic objective.

\paragraph{Structural Entropy and Our Novelty}
A key challenge in GraphRAG is graph partitioning. While structural information theory provides a principled framework for this \cite{li2016structural, rosvall2008maps}, and has been applied in methods like Structural Entropy guided Pooling (SEP) \cite{wu2022structural}, these topological approaches are blind to node semantics. Our work addresses this by proposing \textbf{S²-Entropy}, an objective unifying structural and semantic information. T-Retriever's novelty lies in using S²-Entropy to drive a \textbf{top-down partitioning} of the graph, creating a balanced, multi-resolution index that contrasts sharply with prior bottom-up or heuristic approaches.

\section{Preliminaries}
\label{sec:preliminaries} 
\paragraph{Textual Attributed Graphs.}
We model complex, information-rich data as a \textbf{textual attributed graph}, defined as $\mathcal{G}=(\mathcal{V},\mathcal{E}, \{x_v\}_{v\in \mathcal{V}}, \{x_e\}_{e\in \mathcal{E}})$, where $\mathcal{V} = \{v_1, v_2, \dots, v_n\}$ is the set of $n$ nodes; $\mathcal{E} \subseteq \mathcal{V} \times \mathcal{V}$ is the set of $m$ edges representing relationships between nodes, defining the graph's topology, typically represented by an adjacency matrix $\mathbf{A}$; $x_v \in \mathcal{D}^{L_v}$ denotes the sequential text attribute of node $v \in \mathcal{V}$, where $\mathcal{D}$ is the vocabulary and $L_v$ is the text length; $x_e \in \mathcal{D}^{L_e}$ denotes the sequential text attribute of edge $e \in \mathcal{E}$.
This definition captures attributed graphs where attributes are textual, requiring semantic understanding when analyzed by LLMs.

\paragraph{Retrieval-Augmented Generation (RAG) for Graphs.}
RAG enables users to interact with and query the textual attributed graph $\mathcal{G}$ using natural language, leveraging the power of Large Language Models (LLMs). A graph-based RAG system $\mathcal{M}$ consists of:

\begin{enumerate}
    \item \textbf{Graph Indexer ($\varphi$)}: Processes $\mathcal{G}$ to create an efficient representation for retrieval. This component generates semantic embeddings for nodes ($v$) and edges ($e$) by applying a pre-trained Language Model (LM) to their respective text attributes: $z_v = \text{LM}(x_v) \in \mathbb{R}^d$,
    yielding $d$-dimensional vectors that capture semantic meaning. In our work, $\varphi$ specifically creates an optimized hierarchical encoding tree index over $\mathcal{G}$.
    
    \item \textbf{Graph Retriever ($\psi$)}: Given a natural language query $q$, retrieves the most relevant contextual $G^*$ (e.g., relevant nodes, edges, subgraphs, or information derived from the hierarchical index) from the indexed graph. $\psi(G^*|q, \mathcal{G})$ denotes the process of retrieving context $G^*$.
    
    \item \textbf{Generator ($LLM$)}: An LLM generating the answer $a$ based on the query $q$ and the retrieved graph context $G^*$.
\end{enumerate}

The objective is to generate the optimal answer $a^*$ by maximizing the likelihood:
\begin{equation*}
a^{*} = \arg\max_{a} LLM(a|q, G^{*}) \quad \text{where} \ G^{*} \sim \psi(G|q,\mathcal{G}).
\label{eq:rag_objective_graph}
\end{equation*}

\paragraph{Encoding Tree.}
Given a graph $\mathcal{G}=(\mathcal{V},\mathcal{E}, \{x_v\}_{v\in \mathcal{V}}, \{x_e\}_{e\in \mathcal{E}})$, we define its corresponding Encoding Tree $\mathcal{T}$ as a hierarchical structure with the following properties:

\begin{enumerate}
    \item Each node $\alpha$ in $\mathcal{T}$ is associated with a subset of graph nodes $\mathcal{V}_\alpha \subseteq \mathcal{V}$. The root node $\rho$ corresponds to the entire node set, $\mathcal{V}_\rho = \mathcal{V}$. For a leaf node $\alpha$ at the maximum tree depth $L$, $\mathcal{V}_\alpha$ is a singleton set $\{v\}$ containing exactly one graph node $v \in \mathcal{V}$. Leaf nodes at depths less than $L$ may correspond to empty sets.
    
    \item For every non-leaf node $\alpha$, let $\alpha^{\langle 1 \rangle}, \dots, \alpha^{\langle k_\alpha \rangle}$ be its immediate children, where $k_\alpha$ is the number of children. The parent node of $\alpha$ is denoted $\alpha^-$. The node set $\mathcal{V}_\alpha$ is the disjoint union of the node sets associated with its children: $\mathcal{V}_\alpha = \bigcup_{i=1}^{k_\alpha} \mathcal{V}_{\alpha^{\langle i \rangle}}$.
\end{enumerate}

Consequently, each level of the encoding tree $\mathcal{T}$ implicitly defines a partition of the graph's node set $\mathcal{V}$, with granularity increasing at deeper levels.

\paragraph{Structural Entropy.}
Let $d_v = |\mathcal{N}(v)|$ be the degree of node $v \in \mathcal{V}$ (i.e., the number of its neighbors). The volume of a node subset $\mathcal{S} \subseteq \mathcal{V}$ is defined as the sum of the degrees of the nodes within that set: $\mathrm{Vol}(\mathcal{S}) = \sum_{v \in \mathcal{S}} d_v$. The total volume of the graph is $\mathrm{Vol}(\mathcal{G}) = \sum_{v \in \mathcal{V}} d_v = 2|\mathcal{E}|$.

The structural entropy of graph $\mathcal{G}$ with respect to an encoding tree $\mathcal{T}$ is defined as:
{\small
\begin{equation}
\label{eq:structural_entropy_node_unweighted}
\begin{split}
    H^{\mathcal{T}}(\mathcal{G}) &= \sum_{\alpha\in \mathcal{T}, \alpha\neq \rho} H^{\mathcal{T}}(\mathcal{G}; \alpha), \\
    \text{where} \quad H^{\mathcal{T}}(\mathcal{G}; \alpha) &= -\frac{g_{\alpha}}{\mathrm{Vol}(\mathcal{G})}\log_2\frac{\mathrm{Vol}(\mathcal{V}_{\alpha})}{\mathrm{Vol}(\mathcal{V}_{\alpha^-})}.
\end{split}
\end{equation}
}
Here, $g_{\alpha}$ represents the number of edges connecting nodes within the set $\mathcal{V}_{\alpha}$ to nodes outside $\mathcal{V}_{\alpha}$:
\begin{equation}
    g_\alpha = |\{ (u,v) \in \mathcal{E} \mid u \in \mathcal{V}_\alpha, v \in \mathcal{V} \setminus \mathcal{V}_\alpha \}|.
\end{equation}
$\mathrm{Vol}(\mathcal{V}_{\alpha})$ is the volume of the node set associated with node $\alpha$, and $\alpha^-$ denotes the parent node of $\alpha$.
\section{Methodology}
\label{sec:methodology}

\begin{figure*}[t]
    \centering
    \includegraphics[width=\textwidth]{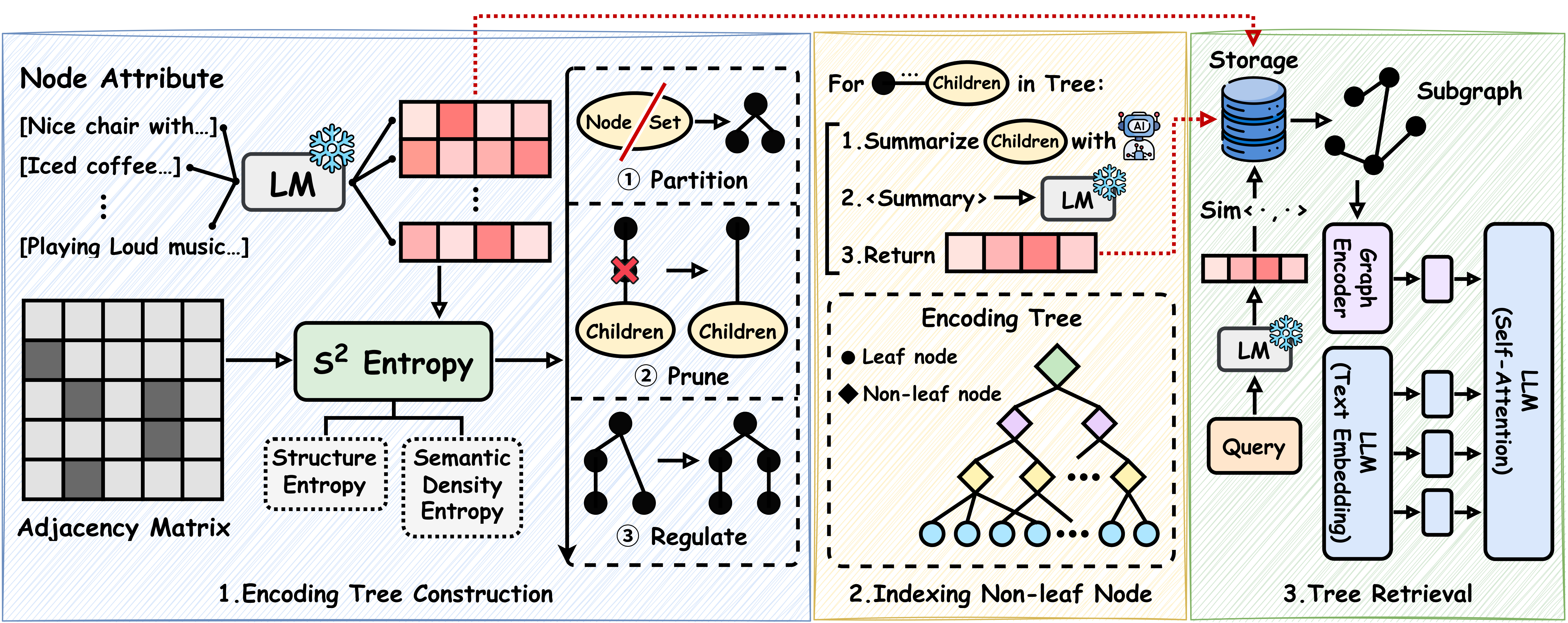}
    \caption{The T-Retriever framework pipeline: (1) Encoding Tree Construction optimizes S²-Entropy (combining structural and semantic information) through partition, prune, and regulate operations; (2) Indexing generates and embeds LLM-based summaries for tree nodes; (3) Tree Retrieval finds relevant nodes, extracts subgraphs, and generates responses using GNN-enhanced LLM prompting.}
    \label{fig:framework}
\end{figure*}

We present T-Retriever, a framework that reformulates attributed graph retrieval through information-theoretic principles. Our approach offers two key innovations: (1) Adaptive Compression Encoding—a globally optimized, learning-free algorithm that constructs hierarchical partitions based on information content rather than predetermined compression quotas; and (2) Semantic-Structural Entropy (S²-Entropy)—a novel metric that unifies topological structure with semantic content to guide partitioning. As shown in Figure~\ref{fig:framework}, T-Retriever constructs an information-theoretically optimal encoding tree, generates hierarchical summaries, and enables efficient multi-resolution retrieval for answering complex graph queries.

\subsection{Semantic-Structural Entropy}
\label{subsec:s2_entropy}

While structural entropy (Eq.~\ref{eq:structural_entropy_node_unweighted}) effectively captures topological information, it overlooks the rich semantics in node attributes $\{x_v\}$. Pure structural partitioning often yields clusters that lack semantic coherence, hampering the RAG system's ability to synthesize relevant information.

To address this limitation, we introduce semantic information into the hierarchical partitioning objective by leveraging node embeddings $z_v = \text{LM}(x_v) \in \mathbb{R}^d$. Rather than computing pairwise distances (which scales as $O(n_\alpha^2)$), we propose a more efficient \textbf{Semantic Density Entropy} that characterizes the embedding distribution in semantic space.

We estimate the probability density function using Kernel Density Estimation:
{\small
\begin{equation}
\label{eq:kde}
\begin{split}
    p(z) &= \frac{1}{n_\alpha} \sum_{v \in \mathcal{V}_\alpha} K_h(z - z_v), \\
    \text{with } K_h(u) &= \frac{1}{(2\pi h^2)^{d/2}} \exp\left(-\frac{\|u\|^2_2}{2h^2}\right).
\end{split}
\end{equation}
}
The semantic density entropy is then defined as: 
{\small
\begin{equation}
\label{eq:semantic_entropy}
H_{sem}(\mathcal{V}_\alpha) = -\frac{1}{n_\alpha} \sum_{v \in \mathcal{V}_\alpha} \log p(z_v),
\end{equation} 
}
where lower values indicate higher semantic coherence.

Our \textbf{Semantic-Structural Entropy (S²-Entropy)} combines structural and semantic components: 
{\small
\begin{equation}
\label{eq:s2_entropy_node}
H_{S^2}(\mathcal{G}; \alpha) = H^{\mathcal{T}}(\mathcal{G}; \alpha) + \lambda H_{sem}(\mathcal{V}_\alpha),
\end{equation}
}
where the hyperparameter $\lambda \ge 0$ balances their relative importance. The total S²-Entropy for tree $\mathcal{T}$ is: $H_{S^2}^{\mathcal{T}}(\mathcal{G}) = \sum_{\alpha \in \mathcal{T}, \alpha \neq \rho} H_{S^2}(\mathcal{G}; \alpha)$.
Minimizing this metric guides partitioning toward clusters that are both structurally well-defined and semantically meaningful, a crucial advancement for effective graph-based RAG systems.

\subsection{Adaptive Compression Encoding}
\label{subsec:entropy_min_partitioning}

The encoding tree organizes the attributed graph hierarchically while preserving essential structural and semantic relationships. By minimizing S²-Entropy, we both reduce structural uncertainty and enhance semantic coherence, creating an optimized knowledge hierarchy.
Our goal is to construct an encoding tree $\mathcal{T}^*$ that minimizes S²-Entropy:
{\small
\begin{equation}
    \mathcal{T}^* = \arg\min_{\forall \mathcal{T}: \text{height}(\mathcal{T}) \leq L} H_{S^2}^{\mathcal{T}}(\mathcal{G}).
\end{equation}
}Since this optimization is generally intractable, we develop an efficient approximation algorithm.  
Unlike the bottom-up approach in Wu et al. \cite{wu2022structural} that iteratively merges nodes, we draw inspiration from Shannon-Fano coding~\cite{connell1973huffman} to develop a top-down recursive partitioning approach. This method offers superior computational efficiency for large-scale attributed graphs and produces partitions that better align with the semantic organization of textual attributes. 
We define three key tree transformation operations:

\begin{definition}[Partition Operation]
$\text{PARTITION}_{\mathcal{T}}(\alpha)$ divides node $\alpha$ with set $\mathcal{V}_\alpha$ into children by solving:
{\small

\begin{equation}
\begin{aligned}
\min_{\mathcal{V}_{\alpha_1}, \mathcal{V}_{\alpha_2}} \quad & H_{S^2}(\mathcal{G}; \alpha_1) + H_{S^2}(\mathcal{G}; \alpha_2) \\
\text{s.t.} \quad & \mathcal{V}_{\alpha_1} \sqcup \mathcal{V}_{\alpha_2} = \mathcal{V}_\alpha, \;\; \mathcal{V}_{\alpha_1}, \mathcal{V}_{\alpha_2} \neq \emptyset.
\end{aligned}
\end{equation}
}
\end{definition}

\begin{definition}[Prune Operation]
The $\text{PRUNE}_{\mathcal{T}}(\alpha)$ operation removes internal node $\alpha$, connecting its children to its parent, $\alpha^-$. The update rule is expressed as:
\[
\alpha^-.\text{children} \leftarrow (\alpha^-.\text{children} \setminus \{\alpha\}) \cup \alpha.\text{children}.
\]
\end{definition}

\begin{definition}[Regulate Operation]
We define the \textbf{Regulate Operation}, denoted $\text{REGULATE}_{\mathcal{T}}(\alpha, \beta)$, which inserts a node $\gamma$ between $\alpha$ and its descendant $\beta$ when their height difference exceeds 1. This process involves two steps:
\begin{enumerate}
    \item Node $\alpha$ adopts $\gamma$ as a new child, replacing $\beta$.
    \[
    \alpha.\text{children} \leftarrow (\alpha.\text{children} \setminus \{\beta\}) \cup \{\gamma\}.
    \]
    \item The new node $\gamma$ adopts $\beta$ as its child.
    \[
    \gamma.\text{children} \leftarrow \{\beta\}.
    \]
\end{enumerate}
\end{definition}

Our algorithm proceeds through three stages:

\begin{enumerate}
\item \textbf{Top-Down Recursive Partitioning}: Starting with the entire graph at the root, we recursively apply the partition operation until reaching either singleton sets or the maximum depth:
{\small
\begin{equation}
\text{PARTITION}_{\mathcal{T}}(\alpha) \text{ for all } \alpha \in \mathcal{T} \text{ where } |\mathcal{V}_\alpha| > 1
\end{equation}
}
\item \textbf{Height Optimization}: If the tree exceeds height $L$, we selectively prune nodes that minimize entropy increase:
{\small
\begin{align}
\alpha = \arg\min_{\alpha \in \mathcal{T} \setminus \{\rho, \text{leaves}\}} \Big\{H_{S^2}^{\mathcal{T}_{\text{PRUNE}(\alpha)}}(\mathcal{G}) \notag\\
\quad - H_{S^2}^{\mathcal{T}}(\mathcal{G})\Big\}
\end{align}
}Unlike conventional methods that impose rigid layer-specific compression quotas, our global optimization approach is guided solely by the desired tree height $L$, enabling better preservation of local graph structures while effectively capturing cross-layer dependencies.

\item \textbf{Structure Regularization}: We ensure proper tree structure by adding intermediate nodes where needed (Regulate Operation), preserving S²-Entropy (Proposition \ref{prop:fill_preserves_entropy}).
\end{enumerate}

\begin{proposition}
\label{prop:fill_preserves_entropy}
For any encoding tree $\mathcal{T}$ and nodes $\alpha, \beta \in \mathcal{T}$ where $\alpha$ is an ancestor of $\beta$ with height difference >1, the Regulate operation preserves S²-Entropy: $H_{S^2}^{\mathcal{T}}(\mathcal{G}) = H_{S^2}^{\mathcal{T}_{\text{REGULATE}(\alpha, \beta)}}(\mathcal{G})$. Proof is provided in Appendix.
\end{proposition}

Our Shannon-Fano inspired approach offers several advantages over bottom-up methods like SEP~\cite{wu2022structural} when dealing with semantic entropy: (1) better semantic coherence preservation across hierarchical levels, as we maintain the global context during partitioning rather than potentially disrupting it through iterative merging; (2) avoiding the quadratic complexity of pairwise node comparisons inherent in bottom-up methods.

\subsection{Indexing with Encoding Tree}
\label{subsec:indexing_with_encoding_tree}

The encoding tree $\mathcal{T}$ organizes graph nodes hierarchically, with each tree node $\alpha \in \mathcal{T}$ representing a subset of graph nodes $\mathcal{V}_\alpha \subseteq \mathcal{V}$. This structure enables multi-resolution knowledge representation, from specific entities at leaf nodes to broader conceptual groups at higher levels.

\textbf{Semantic Content Preparation.}
For leaf nodes representing individual graph nodes, we directly use their original text attributes $x_v$. For non-leaf nodes $\alpha$, we generate summaries using an LLM. Let the set of node attributes in the cluster be $\mathcal{X}_v(\alpha) = \{x_v \mid v \in \mathcal{V}_\alpha\}$ and edge attributes be $\mathcal{X}_e(\alpha) = \{x_e \mid e=(u,v) \in \mathcal{E}, u,v \in \mathcal{V}_\alpha\}$. The summary $S_\alpha$ is then:
{\small
\begin{equation}
S_\alpha = 
\begin{dcases}
    x_v, & \text{if } \alpha \text{ is a leaf with } \mathcal{V}_\alpha = \{v\} \\
    \text{LLM}(\mathcal{X}_v(\alpha), \mathcal{X}_e(\alpha)), & \text{otherwise}
\end{dcases}
\end{equation}
}
These summaries encapsulate key entities, relationships, and concepts in each subtree.

\textbf{Embedding and Index Construction.}
Following Preliminaries section, we use the same language model for encoding both query and node content: $z_\alpha = \text{LM}(S_\alpha) \in \mathbb{R}^d$,
where $\text{LM}(\cdot)$ maps text to a $d$-dimensional embedding space. We organize these embeddings into a multi-level index: $\mathcal{I} = \{(\alpha, z_\alpha, l_\alpha) \mid \alpha \in \mathcal{T}\}$, where $l_\alpha$ denotes the tree level of node $\alpha$. For efficient similarity search, we deploy an approximate nearest neighbor (ANN) structure, enabling logarithmic-time retrieval of relevant tree nodes.

\subsection{Retrieving from Encoding Tree}
\label{subsec:tree_retriever}

We introduce how to achieve efficient query processing utilizing the encoding tree index. It operates through embedding-based retrieval, subgraph extraction, and GNN-augmented generation.

Given a query $q$, we compute its embedding and retrieve the most relevant encoding tree nodes:
{\small
\begin{align}
    z_q &= \text{LM}(q) \in \mathbb{R}^d, \\
    \mathcal{N}_q &= \text{TopK}(\{(\alpha, \text{sim}(z_q, z_\alpha)) \mid \alpha \in \mathcal{T}\}, k).
\end{align}
}
where $\text{sim}(\cdot,\cdot)$ is a similarity function and $k$ controls retrieval volume. This approach treats all encoding tree nodes uniformly regardless of their hierarchical position.

For each retrieved node $\alpha \in \mathcal{N}_q$, we extract its corresponding subgraph:
{\small
\begin{equation}
\begin{split}
    G_\alpha = \big(&\mathcal{V}_\alpha, \mathcal{E}_\alpha, \{x_v\}_{v \in \mathcal{V}_\alpha}, \{x_e\}_{e \in \mathcal{E}_\alpha}\big), \\
    \text{where} \quad \mathcal{E}_\alpha &= \{(u,v) \in \mathcal{E} \mid u,v \in \mathcal{V}_\alpha\}.
\end{split}
\end{equation}
}
The combined retrieval subgraph is: $G_q = \bigcup_{\alpha \in \mathcal{N}_q} G_\alpha$.
Following G-retriever~\cite{he2024g} to integrate structure into the language model, we employ a GNN encoder:
{\small
\begin{align}
    h_g &= \text{POOL}(\text{GNN}_\phi(G_q)) \in \mathbb{R}^{d_g}, \\
    \hat{h}_g &= \text{MLP}_\theta(h_g) \in \mathbb{R}^{d_l}.
\end{align}
}
We textualize the subgraph: $T_q = \text{Textualize}(G_q)$ and generate the final response by:
\begin{equation}
    \text{Response} = \text{LLM}(q, T_q, \hat{h}_g)
\end{equation}

\subsection{The Catalytic Effect of S²-Entropy}
\label{subsec:catalytic_effect}

We now analyze how incorporating semantic information enhances the encoding tree's quality through what we term the "catalytic effect." This effect occurs when semantically similar but structurally distant nodes are grouped together, catalyzing improved overall clustering.

For nodes $u, v \in \mathcal{V}$ with high semantic similarity $\text{sim}(z_u, z_v) > 1-\delta$ but large geodesic distance $d_G(u,v) > \gamma$, pure structural entropy minimization typically separates them into different clusters. However, when S²-Entropy is minimized instead:

\begin{proposition}[Catalytic Effect]
\label{thm:catalytic_effect}
There exists a threshold $\lambda_0 > 0$ such that when $\lambda > \lambda_0$ in S²-Entropy, semantically similar but structurally distant nodes will be placed in the same cluster, catalyzing the inclusion of bridging nodes and yielding lower entropy than any partitioning separating them. Proof in Appendix.
\end{proposition}

The key insight is that this process occurs in two phases: (1) semantic entropy first brings together distant but semantically similar nodes, and (2) structural optimization then incorporates intermediate nodes that form bridges between them. This creates clusters that better reflect semantic relationships and structural aspects of the graph.

This effect is particularly beneficial for retrieval, as information relevant to a query is often distributed across semantically related but structurally distant parts of the graph. By grouping this information together in the encoding tree, our approach enables more comprehensive retrieval.

\section{Experiments}

\subsection{Experimental Setup}
Our framework uses \texttt{Sentence-BERT}~\cite{reimers2019sentence} for encoding and \texttt{Llama-2-7b-chat}~\cite{touvron2023llama} for generation, with Accuracy as the primary metric.

\textbf{Baselines.} To ensure a comprehensive evaluation, we compare against methods representing different RAG philosophies. \textbf{Inference-only}: Standard prompting without retrieval. \textbf{Flat Graph-RAG}: Powerful methods like G-Retriever~\cite{he2024g} and GRAG~\cite{hu2024grag}. \textbf{Hierarchical Graph-RAG}: We adapt state-of-the-art methods to our attributed graph setting. \textbf{RAPTOR}~\cite{sarthi2024raptor} represents a "semantics-first" approach, adapted by applying its text-based clustering to our node attributes. \textbf{ArchRAG}~\cite{wang2025archrag} represents a "structure-first" approach, adapted by using its community detection method on our graph topology.

\subsection{Main Results}

\begin{table*}[t]
\centering
\resizebox{0.9\textwidth}{!}{%
\begin{tabular}{llccc}
\toprule
   \textbf{Setting} & \textbf{Method} & \textbf{SceneGraphs} & \textbf{WebQSP} & \textbf{BookGraphs} \\
   & & \textbf{(Avg. 19 nodes)} & \textbf{(Avg. 1,371 nodes)} & \textbf{(Avg. 76,875 nodes)} \\
\midrule
\multirow{4}{*}{\textbf{Inference-only}} 
    & Zero-shot & 0.4012$_{\pm 0.0388}$ & 0.4118$_{\pm 0.0183}$ & 0.3169$_{\pm 0.0145}$ \\
    & Zero-CoT & 0.5217$_{\pm 0.0098}$ & 0.5104$_{\pm 0.0228}$ & 0.3785$_{\pm 0.0091}$ \\
    & CoT-BAG & 0.5587$_{\pm 0.0198}$ & 0.4029$_{\pm 0.0184}$ & 0.3843$_{\pm 0.0372}$ \\
    & KAPING & 0.4598$_{\pm 0.0187}$ & 0.5374$_{\pm 0.0269}$ & 0.4298$_{\pm 0.0129}$ \\
\midrule
\multirow{4}{*}{\textbf{Flat Graph-RAG}} 
    & G-Retriever w/ PT & 0.8102$_{\pm 0.0311}$ & 0.6772$_{\pm 0.0182}$ & 0.6603$_{\pm 0.0211}$ \\
    & G-Retriever w/ LoRA & \underline{0.8311$_{\pm 0.0199}$} & 0.7186$_{\pm 0.0394}$ & 0.6715$_{\pm 0.0385}$ \\
    & GRAG w/ PT & 0.7988$_{\pm 0.0375}$ & 0.7228$_{\pm 0.0283}$ & 0.6689$_{\pm 0.0392}$ \\
    & GRAG w/ LoRA & 0.8017$_{\pm 0.0486}$ & {0.7254$_{\pm 0.0477}$} & {0.6738$_{\pm 0.0573}$} \\
\midrule
\multirow{2}{*}{\textbf{Hierarchical Graph-RAG}}
    & RAPTOR (Semantics-first) & 0.7995$_{\pm 0.0215}$ & 0.7114$_{\pm 0.0250}$ & 0.6819$_{\pm 0.0288}$ \\
    & ArchRAG (Structure-first) & {0.8212$_{\pm 0.0180}$} & \underline{0.7533$_{\pm 0.0291}$} & \underline{0.7108$_{\pm 0.0345}$} \\
\midrule
\multirow{2}{*}{\textbf{Ours}} 
    & \textbf{T-Retriever} & \textbf{0.8507}$_{\pm \textbf{0.0121}}$ & \textbf{0.7715}$_{\pm \textbf{0.0387}}$ & \textbf{0.7579}$_{\pm \textbf{0.0183}}$ \\
    & \textit{Improvement} & $\uparrow$ 2.36\% & $\uparrow$ 2.42\% & $\uparrow$ 6.63\% \\
\bottomrule
\end{tabular}
}
\caption{Performance comparison across SceneGraphs, WebQSP, and BookGraphs datasets. \textbf{Boldface} denotes the highest score, and \underline{underline} indicates the best result among baselines.}
\label{tab:results}
\end{table*}

Table \ref{tab:results} shows that T-Retriever consistently outperforms all baselines. Our key findings are:

\begin{itemize}
    \item \textbf{Value of Hierarchy is Nuanced:} While hierarchical indexing is a promising direction, its effectiveness depends on the strategy. The structure-first ArchRAG shows significant gains over flat methods on complex graphs (WebQSP, BookGraphs). However, the semantics-first RAPTOR, which ignores graph topology, struggles on these tasks and is sometimes outperformed by strong flat baselines like GRAG.
    
    \item \textbf{Joint Optimization is Key:} T-Retriever achieves the best performance by outperforming all competitors, including the strong structure-aware ArchRAG. This validates our core hypothesis: by jointly optimizing structure and semantics via S²-Entropy, our method creates a more effective knowledge hierarchy than approaches that prioritize one aspect over the other.
    
    \item \textbf{Gains Scale with Complexity:} The performance advantage of T-Retriever is most pronounced on the largest dataset, BookGraphs ($\uparrow$6.63\%). This pattern confirms that as graph size and complexity increase, the benefits of our principled partitioning become increasingly significant. For smaller graphs like SceneGraphs, the limited tree depth causes T-Retriever to more closely approximate other hierarchical methods, resulting in a smaller performance gap.
\end{itemize}

\subsection{Hyperparameter Analysis}
We conduct sensitivity analysis of three key hyperparameters: (1) the number of encoding tree layers $L$, (2) the number of retrieved subgraphs $k$, and (3) the S²-Entropy weighting factor $\lambda$ and the KDE bandwidth h.

\begin{figure*}[t]
    \centering
    \begin{subfigure}{0.25\textwidth}
        \centering
        \includegraphics[width=\linewidth]{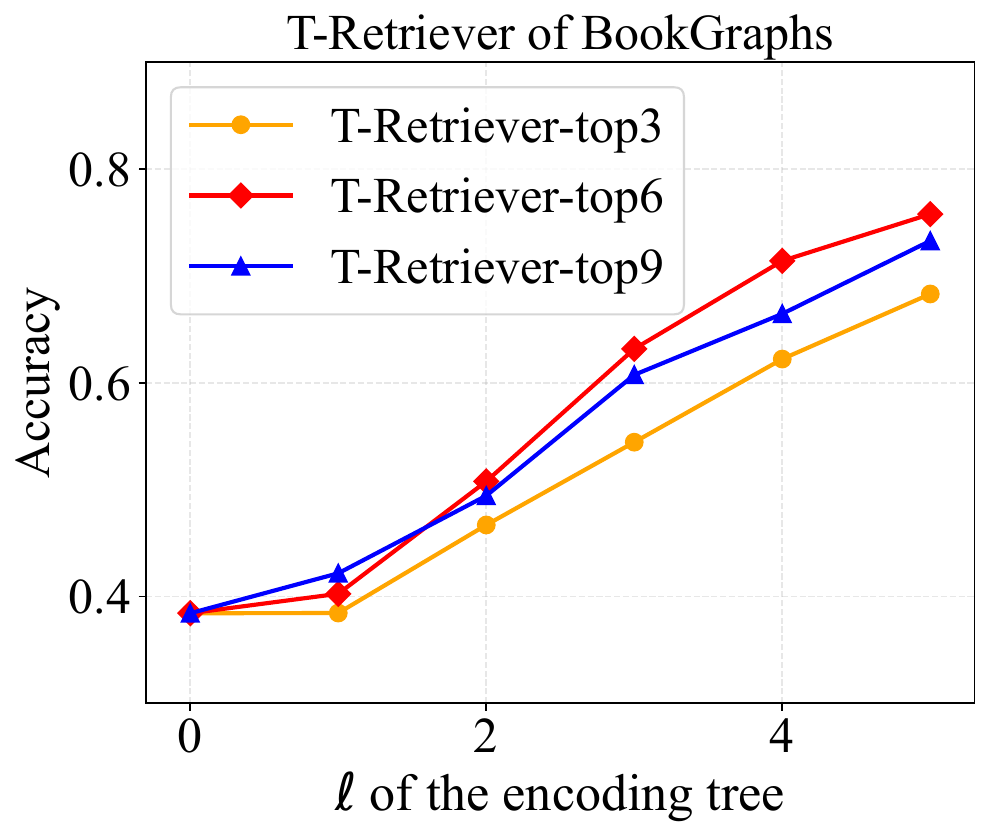}
        \caption{BookGraphs Analysis}
        \label{fig:sub_book}
    \end{subfigure}
    \hfill 
    \begin{subfigure}{0.25\textwidth}
        \centering
        \includegraphics[width=\linewidth]{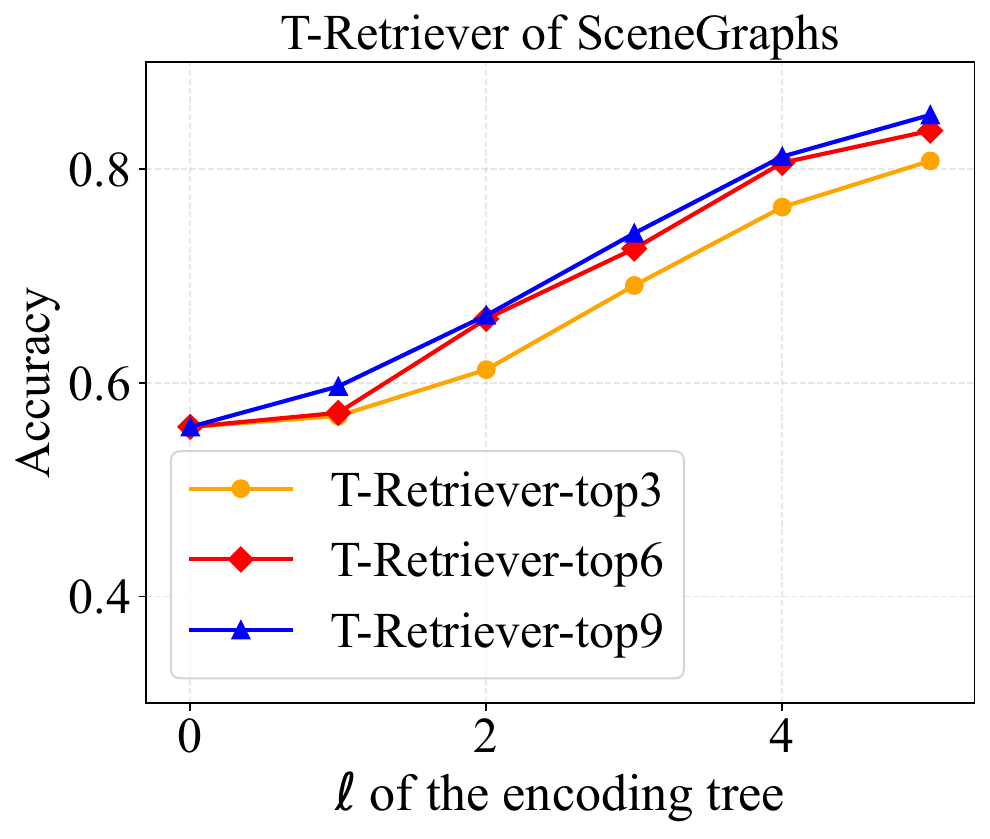}
        \caption{SceneGraphs Analysis}
        \label{fig:sub_scene}
    \end{subfigure}
    \hfill 
    \begin{subfigure}{0.25\textwidth}
        \centering
        \includegraphics[width=\linewidth]{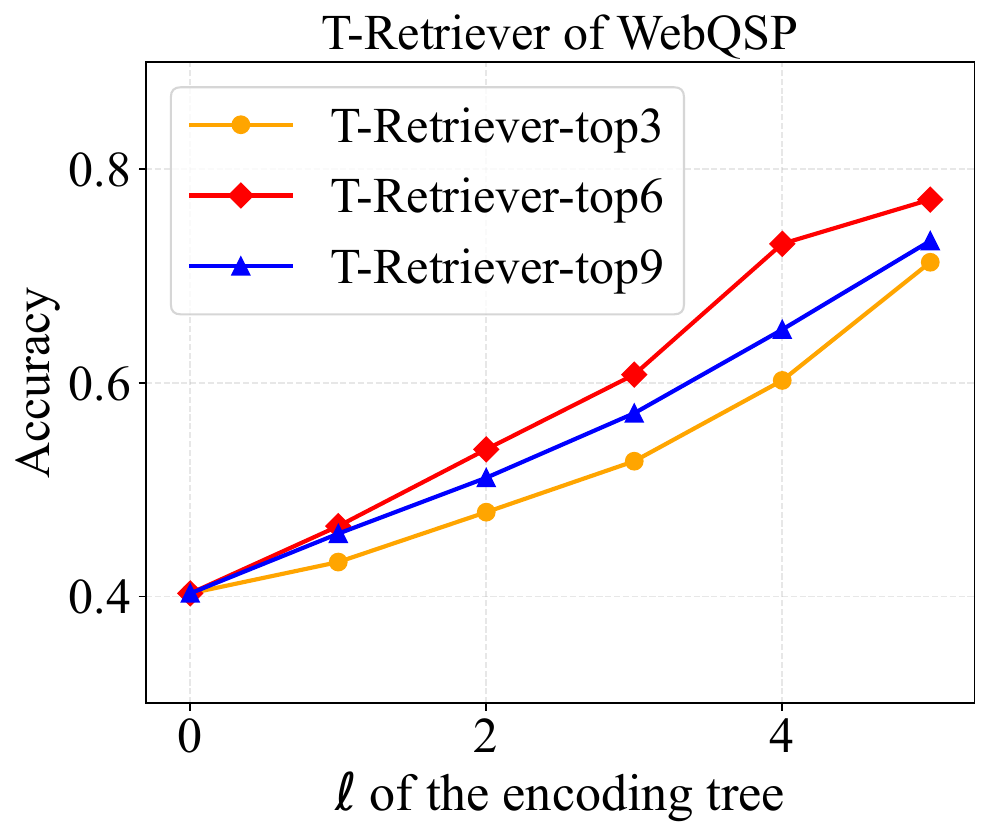}
        \caption{WebQSP Analysis}
        \label{fig:sub_webqsp}
    \end{subfigure}
    \hfill 
    \begin{subfigure}{0.22\textwidth}
        \centering
        \includegraphics[width=\linewidth]{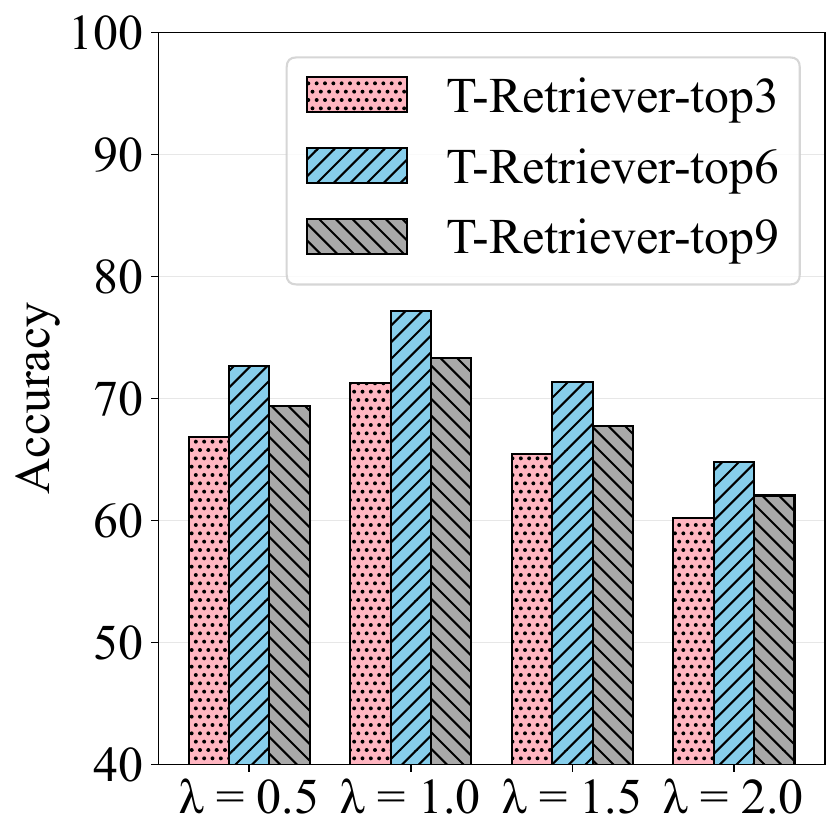}
        \caption{Impact of weight $\lambda$}
        \label{fig:sub_lambda}
    \end{subfigure}
    
    \caption{Sensitivity analysis of key hyperparameters. (a-c) Impact of encoding tree layers $L$ and retrieved subgraphs $k$ across different datasets. (d) Impact of the entropy weighting factor $\lambda$ on WebQSP.}

    \label{fig:hyperparameter_analysis_combined}
\end{figure*}

\begin{itemize}
    \item \textbf{Number of Encoding Tree Layers $L$}: 
    Figure~\ref{fig:hyperparameter_analysis_combined}(a-c) demonstrates that \texttt{T-Retriever}'s accuracy increases with encoding tree depth. Deeper hierarchy better capture the multi-resolution nature of attributed graphs, allowing the encoding tree to represent both fine-grained local structure and broader semantic relationships, creating increasingly optimized information partitioning.
    
    \item \textbf{Number of Retrieved Subgraphs $k$}:
    As shown in Figure~\ref{fig:hyperparameter_analysis_combined}(a-c), accuracy generally improves with more retrieved subgraphs, providing wider contextual information for the LLM. However, for larger graphs, excessive retrieval may introduce noise and increase LLM processing complexity, necessitating a balance between retrieval effectiveness and efficiency. Our experiments identify $k=6$ as the typical optimal value.

    \item \textbf{S²-Entropy Hyperparameters ($\lambda$ and $h$).} We analyzed the hyperparameters governing S²-Entropy. As shown for WebQSP in Figure~\ref{fig:hyperparameter_analysis_combined}(d), performance peaks around $\lambda=1.0$. Our validation across datasets confirms that performance is most robust when $\lambda$ is in the [1.0, 1.5] range, suggesting a near-equal balance is broadly effective. The critical KDE bandwidth $h$ (Eq. 4) was determined for each dataset via a principled, cross-validated grid search, which prevents performance loss from overfitting or oversmoothing the semantic distribution.

\end{itemize}

\begin{figure}[t]
    \centering
    \includegraphics[width=\columnwidth]{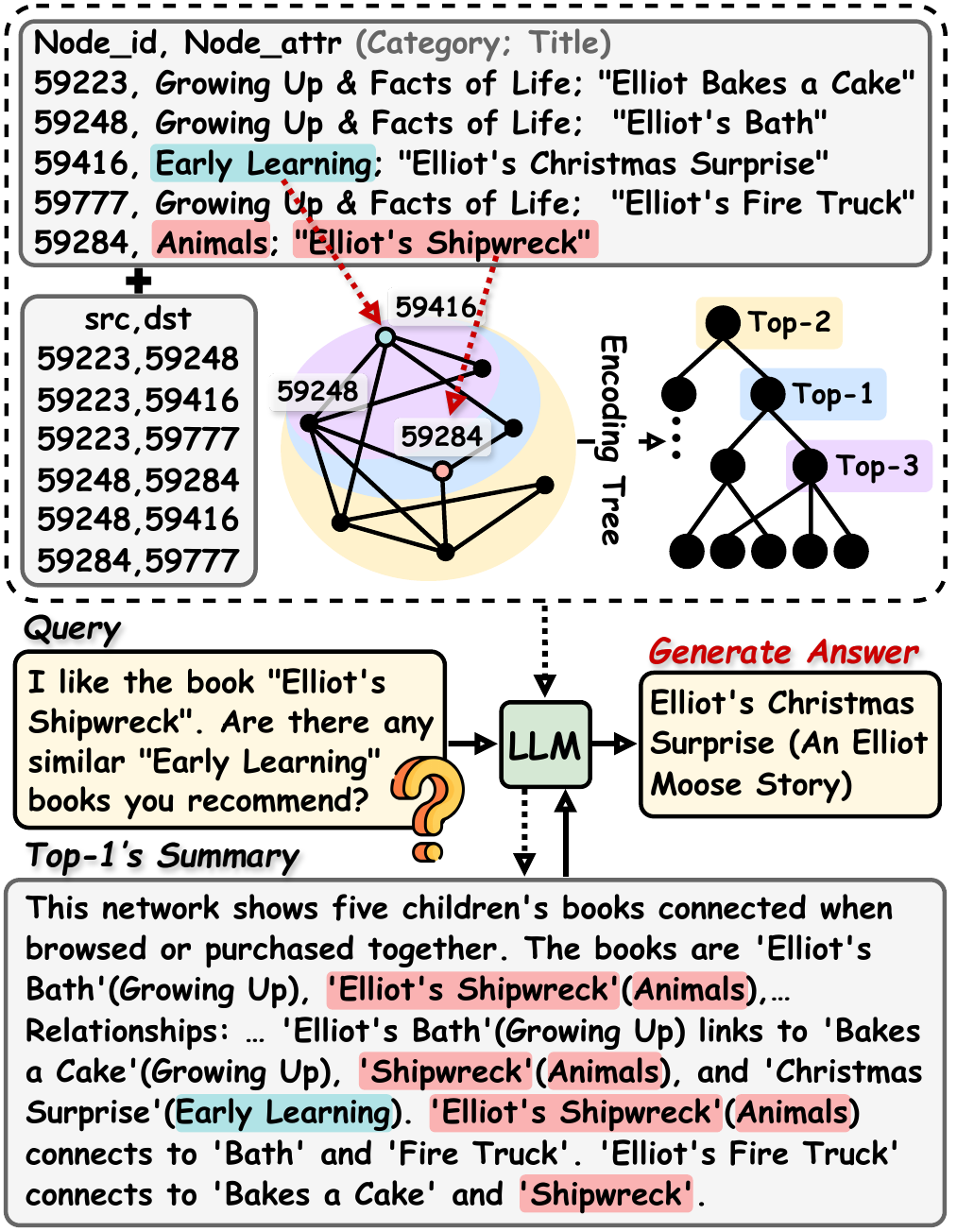}
    \caption{Case study visualization from BookGraphs.}
    \label{fig:case}
\end{figure}

To empirically validate \texttt{T-Retriever}'s effectiveness, we conduct analyses on the WebQSP benchmark. As depicted in Figure \ref{fig:case}, \texttt{T-Retriever} operates in three phases: (1) constructing a hierarchical encoding tree via S²-Entropy optimization, (2) retrieving subgraphs closely aligned with the query in semantic and structural aspects, and (3) converting these subgraphs into standardized formats for LLM-based answer generation. The joint S²-Entropy optimization ensures structural integrity and optimal semantic alignment with the query, boosting accuracy and efficiency.

\subsection{Ablation Study}
To rigorously evaluate the contribution of individual components within the Semantic-Structural (S²) Entropy, we conducted a comprehensive ablation study on the WebQSP dataset. We systematically compared three configurations: (1) Semantic-Only entropy optimization, (2) Structural-Only entropy optimization, and (3) our proposed joint S²-Entropy integration. 

\begin{table}[t]
\centering
\small
\resizebox{0.45\textwidth}{!}{%
\begin{tabular*}{\columnwidth}{@{\extracolsep{\fill}}lccc@{}}
\toprule
\textbf{Configuration} & \textbf{Acc.} & \textbf{F1} & \textbf{Rec.} \\
\midrule
Semantic-Only & \makecell{0.6319 \\ \scriptsize($\downarrow$18.09\%)} & \makecell{0.4055 \\ \scriptsize($\downarrow$21.69\%)} & \makecell{0.3981 \\ \scriptsize($\downarrow$24.67\%)} \\
\addlinespace 
Structural-Only & \makecell{0.7154 \\ \scriptsize($\downarrow$7.27\%)} & \makecell{0.4649 \\ \scriptsize($\downarrow$10.22\%)} & \makecell{0.4780 \\ \scriptsize($\downarrow$9.56\%)} \\
\addlinespace
\makecell[l]{S²-Entropy} & \textbf{0.7715} & \textbf{0.5178} & \textbf{0.5285} \\
\bottomrule
\end{tabular*}
}
\caption{Ablation study comparing different entropy configurations on WebQSP.}
\label{tab:ablation}
\end{table}

The results in Table \ref{tab:ablation} reveal several critical insights. First, structural entropy demonstrates substantially greater influence on retrieval performance than semantic entropy alone, aligning with graph information theory principles suggesting that topological structure provides a fundamental scaffold for information organization. Second, Semantic-Only captures similarity but fails to preserve structural coherence, while Structural-Only lacks semantic relevance due to discarded node attributes. Finally, S²-Entropy achieves balanced performance by simultaneously optimizing both aspects, yielding retrievals that are both structurally cohesive and semantically relevant.

\subsection{Efficiency Evaluation}

\begin{table}[t]
\centering
\small
\resizebox{0.5\textwidth}{!}{%
\begin{tabular}{lccc}
\toprule
\textbf{Metric} & \textbf{BookGraphs} & \textbf{SceneGraphs} & \textbf{WebQSP} \\
\midrule
\multicolumn{4}{l}{\textit{Before Retrieval (Avg.)}} \\
\quad \# Tokens & 5,377,476 & 1,396 & 100,627 \\
\quad \# Nodes  & 76,876    & 19    & 1,371 \\
\midrule
\multicolumn{4}{l}{\textit{After G-Retriever (Avg.)}} \\
\quad \# Tokens & 3,973.4 & 346.0 & 721.0 \\
\quad \# Nodes  & 213     & 9     & 39 \\
\midrule
\multicolumn{4}{l}{\textit{After T-Retriever (Avg.)}} \\
\quad \# Tokens & \textbf{1,487.6} & \textbf{123.1} & \textbf{114.2} \\
\quad \textit{Reduction} & \textit{($\downarrow$62.56\%)} & \textit{($\downarrow$64.42\%)} & \textit{($\downarrow$84.16\%)} \\
\quad \# Nodes  & \textbf{33} & \textbf{7} & \textbf{26} \\
\quad \textit{Reduction} & \textit{($\downarrow$84.51\%)} & \textit{($\downarrow$22.22\%)} & \textit{($\downarrow$33.33\%)} \\
\bottomrule
\end{tabular}
}
\caption{Efficiency comparison of graph processing methods across three benchmark datasets.}
\label{tab:efficiency}
\end{table}

\begin{table}[h]
\centering
\small
\begin{tabular}{lrr}
\toprule
\textbf{Preprocessing Stage} & \textbf{WebQSP} & \textbf{BookGraphs} \\
\midrule
Node Embedding      & 50.3 sec         & 18.7 min \\
S²-Entropy Partitioning & 5.7 min          & 4.2 hours \\
Summarization \& Indexing & 9.0 min          & 3.1 hours \\
\midrule
\textbf{Total Time} & \textbf{\textasciitilde{}14.8 min} & \textbf{\textasciitilde{}7.3 hours} \\
\bottomrule
\end{tabular}

\caption{Wall-clock time for T-Retriever's one-time, offline preprocessing stages on a single NVIDIA A100 GPU.}
\label{tab:preprocessing_time}
\end{table}

Our analysis highlights two key aspects: online retrieval benefits and offline preprocessing costs.
First, as shown in Table~\ref{tab:efficiency}, T-Retriever significantly reduces the context size of LLMs compared to G-Retriever, cutting token counts by up to 84.16\%. This demonstrates its superior online efficiency.
Second, to ensure a thorough assessment of practical scalability, we report the one-time offline costs in Table~\ref{tab:preprocessing_time}. The total indexing time for the large 77k-node BookGraphs dataset is approximately 7.3 hours, a practical one-time investment for a high-quality, reusable index. This demonstrates that the substantial online efficiency gains are achieved with a reasonable and transparent offline computational budget.

\section{Conclusions}
\label{sec:conclusion}
We introduced T-Retriever, a framework that reformulates attributed graph retrieval via an information-theoretically optimal encoding tree. By minimizing our proposed S²-Entropy—a novel metric unifying semantic content and topological structure—T-Retriever builds a multi-resolution knowledge hierarchy that is both structurally and semantically coherent. A key finding is that this superior knowledge organization allows T-Retriever to outperform even fine-tuned baselines without requiring any parameter updates, highlighting the efficiency and power of our approach for complex graph reasoning. Consequently, experiments across diverse benchmarks confirm that T-Retriever sets a new state-of-the-art.

\section{Acknowledgments}
This work was supported by The Disciplinary Breakthrough Project of Ministry of Education (MOE, No.00975101), NSFC (No.6250072448, No.62272466, U24A20233), and Big Data and Responsible Artificial Intelligence for National Governance, Renmin University of China.

\nocite{*}
\bibliography{main}

\end{document}